\documentclass[journal]{IEEEtran}
\IEEEoverridecommandlockouts
\usepackage{amsmath,amssymb,amsfonts}
\usepackage{graphicx}
\usepackage{textcomp}
\usepackage{xcolor}
\usepackage{colortbl}
\usepackage{bm}
\usepackage[switch]{lineno}
\usepackage{booktabs}
\usepackage{amsmath}
\usepackage{amssymb}
\usepackage{amsthm}
\usepackage{mathrsfs}
\usepackage{enumerate}
\usepackage{multirow}
\usepackage{color}
\usepackage{threeparttable}
\usepackage{subfigure}
\usepackage{makecell}
\usepackage[utf8]{inputenc}
\usepackage[square, comma, sort&compress, numbers]{natbib}
\usepackage[pagebackref=false,breaklinks=true,colorlinks,bookmarks=false]{hyperref}
\usepackage{times}
\usepackage{array}
\usepackage{verbatim}
\usepackage{algorithm}
\usepackage{bbding}
\usepackage{algpseudocode}
\usepackage{lettrine}
\def\BibTeX{{\rm B\kern-.05em{\sc i\kern-.025em b}\kern-.08em
    T\kern-.1667em\lower.7ex\hbox{E}\kern-.125emX}}
\begin{document}

\title{Photovoltaic Defect Image Generator with Boundary Alignment Smoothing Constraint for Domain Shift Mitigation\\ 
}

\author{
Dongying Li,
Binyi Su,
Hua Zhang,
Yong Li,
Haiyong Chen
      
\thanks{This work was supported in part by the 
    National Natural Science Foundation of China under Grant 62473127, National Natural Science Foundation of China Joint Foundation Program U21A20482, National Key Research and Development Program of China under Grant 2024YFB3310904 and 2021YFB3100800, Central Government Guides Local Science and Technology Development Fund Project 246Z1602G, Major Science and Technology Support Program of Hebei Province 242Q4302Z, Natural Science Foundation of China under Grant 62272018 and 62072454, Beijing Natural Science Foundation under Grant 4202084, Basal Research Fund of Central Public Research Institute of China under Grant 20212701 and 246Z4306G, Shijiazhuang Science and Technology Cooperation Project under Grant SJZZXC24009, Tianjin Municipal Education Commission Scientific Research Project under Grant 2024KJ151.(\textit{Corresponding authors: Haiyong Chen.})
        
        D. Li and H. Chen are with the School of Artificial Intelligence and Data Science, Hebei University of Technology, Tianjin 300401, China (e-mail: 202312801005@stu.hebut.edu.cn, haiyong.chen@hebut.edu.cn). 
        
        B. Su is with the School of Artificial Intelligence and Data Science, Hebei University of Technology, Tianjin 300401, China, and also with China Xiongan Group Digital City Technology Company Ltd., Hebei 070001, China(e-mail:subinyi@vip.qq.com).
        
        H. Zhang is with the State Key Laboratory of Information Security, Institute of Information Engineering, Chinese Academy of Sciences, Beijing 100093, China (e-mail: zhanghua@iie.ac.cn).
		
	Y. Li is with the school of Instrumentation and Optoelectronic Engineering, Beihang University, Beijing 100191, China (e-mail: by2017335@buaa.edu.cn)}
         }

\markboth{Journal of \LaTeX\ Class Files,~Vol.~14, No.~8, August~2021}%
{Shell \MakeLowercase{\textit{et al.}}: A Sample Article Using IEEEtran.cls for IEEE Journals}

\maketitle

\begin{abstract}
Accurate defect detection of photovoltaic (PV) cells is critical for ensuring quality and efficiency in intelligent PV manufacturing systems. However, the scarcity of rich defect data poses substantial challenges for effective model training. While existing methods have explored generative models to augment datasets, they often suffer from instability, limited diversity, and domain shifts. 
To address these issues, we propose PDIG, a \underline{P}hotovoltaic \underline{D}efect \underline{I}mage \underline{G}enerator based on Stable Diffusion (SD). PDIG leverages the strong priors learned from large-scale datasets to enhance generation quality under limited data. Specifically, we introduce a \underline{S}emantic \underline{C}oncept \underline{E}mbedding (SCE) module that incorporates text-conditioned priors to capture the relational concepts between defect types and their appearances. To further enrich the domain distribution, we design a \underline{L}ightweight \underline{I}ndustrial \underline{S}tyle \underline{A}daptor (LISA), which injects industrial defect characteristics into the SD model through cross-disentangled attention.
At inference, we propose a \underline{T}ext-\underline{I}mage \underline{D}ual-\underline{S}pace \underline{C}onstraints (TIDSC) module, enforcing the quality of generated images via positional consistency and spatial smoothing alignment. Extensive experiments demonstrate that PDIG achieves superior realism and diversity compared to state-of-the-art methods. Specifically, our approach improves Frechet Inception Distance (FID) by 19.16 points over the second-best method and significantly enhances the performance of downstream defect detection tasks.

\end{abstract}


\section{Introduction}
\IEEEPARstart{P}{hotovoltaic} (PV) cells are prone to various defects during different production processes, such as microcracks, scratches, and spots, which severely affect the lifespan and power generation efficiency of the PV cells~\cite{dhimish2021investigating}. To ensure the quality and safety of industrial products, comprehensive detection of defects on PV EL images is imperative.

In recent years, deep learning-based object detection has become a cornerstone in the field of computer vision. Su et al.~\cite{su2020deep,su2021baf} and Zhao et al.~\cite{zhao2023sncf} designed a multi-scale attention mechanism based on CNN to refine multi-scale features, thus improving the performance of classification and detection.
\begin{figure}[t]
\centering
\includegraphics[width=1\linewidth]{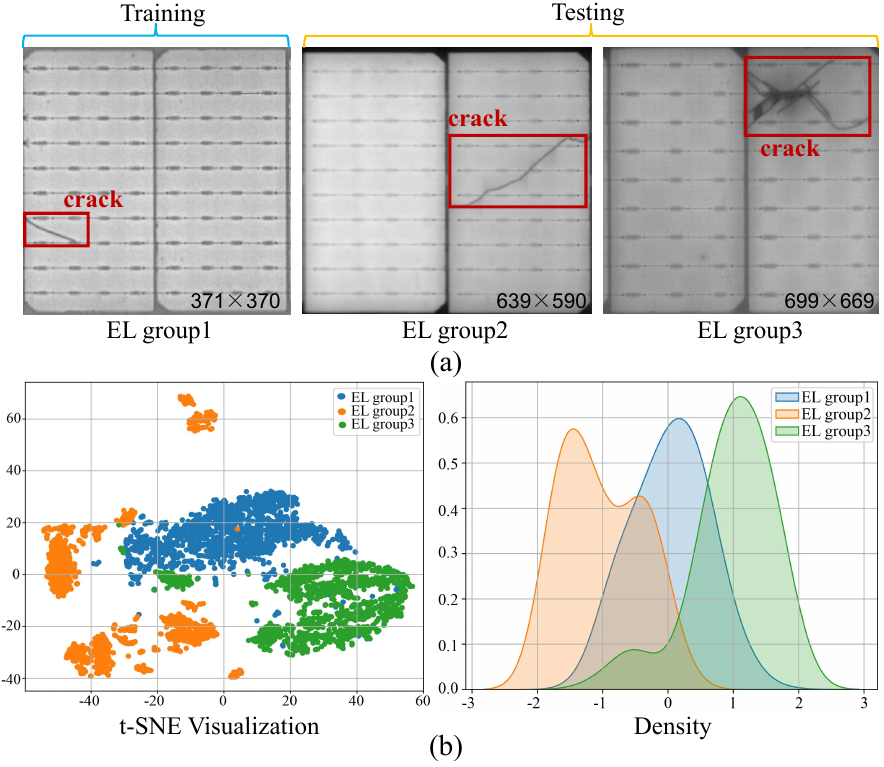}
\caption{The published ELES dataset exhibits issues of domain shift and instance shift. (a) Domain shift: variations exist in the resolution, background brightness, and grid line spacing of defect images across different production lines; Instance shift: the distribution of instances in the training dataset is inconsistent with that in the test data. For example, cracks in the training set appear as fine, linear defects, whereas those in the testing set are characterized by large-area damage with inconsistent shapes. (b) t-SNE dimensionality reduction and probability distribution visualization on the data from different production lines reveal data distribution deviations caused by endogenous shifts.}
\label{fig:1}
\end{figure}
However, these methods only consider scenarios of single production lines, ignoring the complex background styles and instance shift across multiple production lines. Fig. \ref{fig:1} (a) shows the domain shift and instance shift issues present in datasets from different production lines. As shown in Fig. \ref{fig:1} (b), t-SNE and probability density graph verify that there is an obvious shift problem between the datasets.
Therefore, the Shift Suppression Network~\cite{zhao2023ssn} addresses the problem of endogenous displacement in PV defect detection by learning enhanced feature representations through style alignment and cross-layer interaction.
However, in actual production processes, as the diversity of defects in the images increases, the shift problems become more pronounced~\cite{wang2024unveiling}. 
Neural networks are less effective due to the limitations of the training data. Thus, employing image generation techniques to enhance the quantity and diversity of defect data is a key strategy to mitigate the shift problem.
\begin{figure}[t]
\centering
\includegraphics[width=1\linewidth]{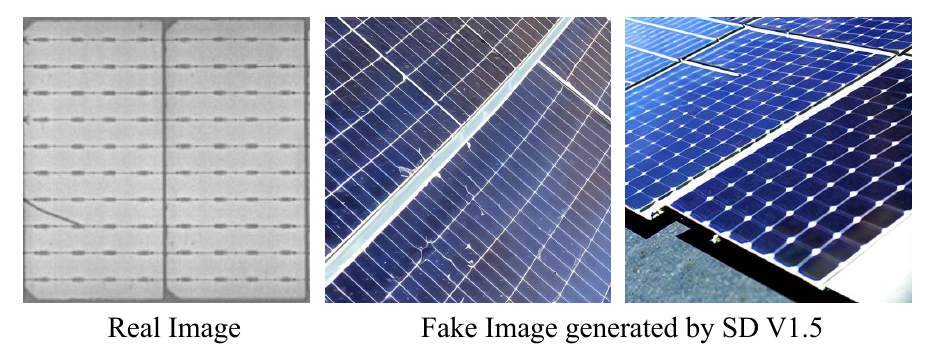}
\caption{The images generated by the SD V1.5 model with the prompt “PV defect images based on EL imaging”.}
\label{fig:2}
\end{figure}
The traditional method of expanding data based on image processing technology~\cite{shorten2019survey} is limited to the feature space range of the known data. Therefore, high-quality images can be generated by employing the generative adversarial network (GAN)~\cite{creswell2018generative} and its variants.
Defect-GAN~\cite{zhang2021defect} adopts a novel component layer-based architecture to simulate the random changes of defects, which synthesizes various types of defects with excellent diversity and fidelity, meanwhile, improving the performance of the defect detection network. Inspired by the principle of image-to-image conversion, TransP2P~\cite{zhao2023defect} incorporates the advantages of transformer global feature perception and U-Net extracting local detail features to convert defect-free images into defect images. However, training these generative models from scratch is challenging when image samples are insufficient, which often leads to distortion of the generated samples and a lack of authenticity.

Recently, image generation based on the diffusion model~\cite{cao2024survey, shen2024imagpose, shen2025imagdressing} has shown impressive capabilities, as it can embed a vast number of image distributions and generate samples with high fidelity and diversity. Among them, the Stable Diffusion model (SD)~\cite{rombach2022high} has adopted a more stable, controllable, and efficient approach to generating high-quality images. This has led to significant advancements in the quality, speed, and cost of image generation. However, the SD model contains richly diverse and abstracted textual descriptions, which limit users to concepts that the network has already been trained on. As shown in Fig. \ref{fig:2}, when the prompt “PV defect images based on EL imaging” is input, the gap between the image generated by the SD model and the target image is significant. This is because the SD model does not understand the concept of PV EL imaging, and thus cannot generate the defect morphology of actual PV quality inspection images. Meanwhile, due to the limitations of data and computing power, it is very difficult to retrain the SD model.

Several adapter-based methods have been proposed to enhance the controllability of text-to-image (T2I) generation. These approaches typically freeze the base Stable Diffusion (SD) model and introduce lightweight trainable parameters to inject external control signals~\cite{10528891}. ControlNet~\cite{zhang2023adding} supports auxiliary conditions such as edge maps, segmentation maps, and depth. IP-Adapter~\cite{ye2023ip} facilitates image enhancement and restoration tasks, while T2I-Adapter~\cite{mou2024t2i} enables fine-grained control over color and structure.
However, most T2I methods struggle with spatial alignment and complex prompt understanding in industrial scenarios, such as defect image generation. Recent studies have highlighted the importance of structured conditioning: IMAGGarment~\cite{shen2025imaggarment} achieves fine-grained garment control through a diffusion-based framework, and RCDM~\cite{shen2025long} model long-term consistency for talking face generation using motion priors. Motivated by these advances, we propose a targeted conditional framework to better handle spatial and descriptive control in industrial settings.

In this paper, we propose a Photovoltaic Defect Image Generator (PDIG) that employs the strong prior information of the SD model learned from large-scale datasets to enhance the authenticity of the generation under few-shot training data. First, PDIG uses 3-5 industrial defect types and images to enrich the text-embedded priors, capturing the specific relational concepts between defect types and their appearances. Secondly, the LISA is developed to embed the PV EL defect features into the pre-trained cultural graph diffusion model to enhance the image generation domain distribution. Finally, at inference, the TIDSC is proposed to ensure the position and space consistency of the generated defective image. Extensive experiments show that our model significantly outperforms state-of-the-art methods in terms of generated authenticity and diversity, and effectively improves the performance of downstream defect object detection tasks.

The main contributions are summarized as follows:
\begin{itemize}
\item[$\bullet$] To mitigate domain shift in industrial defect detection, we propose the PDIG, a diffusion-based method that augments diverse and realistic defect samples, effectively bridging cross-domain distribution gaps and enhancing model generalization. 
\item[$\bullet$] We propose a novel LISA module that embeds the image features of industrial defects into the SD model, thereby enabling the model to perform multimodal learning of defect features. Meanwhile, to reduce the cost of manual annotation, we propose a TIDSC module to achieve spatial localization generation, providing annotation priors for subsequent defect detection applications.
\item[$\bullet$] The experimental results show that, compared with previous defect generation methods, this method can significantly improve the diversity and realism of defect image generation. In addition, improved domain-adaptive performance for defect detection.
\end{itemize}

\section{Related Work}\label{sec:rw}

\subsection{Defect detection method based on image generation} 
In recent years, GANs and their variants have been widely applied in the field of defect image generation.~\cite{niu2021region,9439889,zhang2024dp,9623476}. To address the class imbalance issue, AdaBalGAN~\cite{wang2019adabalgan} combines the CGAN~\cite{mirza2014conditional} with an adaptive generation controller to produce high-fidelity images of specified categories, significantly enhancing the accuracy and stability of defect recognition. 
Faced with the challenge of insufficient sample diversity, reference ~\cite{tian2019detection} employed the CycleGAN to augment data, thereby enhancing the diversity of the training dataset. Furthermore, SDGAN~\cite{niu2020defect} introduces a D2 adversarial loss to increase diversity. 
To further enhance the quality of image generation, DG2GAN~\cite{deng2024dg2gan} incorporates a cycle-consistent loss, a DJS-optimized discriminator loss, and a DG2 adversarial loss, optimizing the image feature distribution to generate high-quality and diverse defect images. However, these methods inherently tend to merely imitate existing content, leading to mode collapse. In comparison, we employ the SD model as our baseline model. It offers greater creative freedom and better balances image quality with computational cost, thus avoiding mode collapse.

\subsection{Denoising Diffusion Probabilistic Model}
Since their introduction to image generation in 2015, diffusion models~\cite{sohl2015deep} have emerged as a prominent research direction due to their robust generative capabilities and broad application potential. These models perturb data distributions through a forward diffusion process and subsequently learn an inverse diffusion process to recover the original data distribution, resulting in a highly flexible and computationally efficient generative framework.
A notable method in this area is the Denoising Diffusion Probabilistic Model (DDPM)~\cite{ho2020denoising}, which leverages iterative gradient updates combined with Gaussian noise to generate high-fidelity samples, approximating the underlying data manifold through annealed Langevin dynamics. To further improve the sampling efficiency of DDPM, DDIM~\cite{song2020denoising} introduces a non-Markovian diffusion process and deterministic sampling strategy, significantly reducing inference time and computation steps.
To enhance controllability during generation, ILVR~\cite{choi2021ilvr} guides DDPM to produce high-quality images conditioned on reference signals, although it still operates purely in pixel space, resulting in high computational overhead and limited support for multimodal conditioning.
Inspired by these advances, we propose the LISA module, which learns image features by training only a small number of parameters. Our method achieves a thorough integration of image and text modalities, enabling rich multimodal inputs while maintaining high computational efficiency.

\subsection{Latent Diffusion Model}
Latent Diffusion Models (LDMs) were introduced by Rombach et al.~\cite{rombach2022high}, proposing to learn the feature distribution through a diffusion process in the latent space rather than the pixel space. Compared to conventional pixel-space diffusion models, LDMs capture higher-level features and semantic information more efficiently. Notably, LDMs incorporate a conditional encoder that injects control conditions (e.g., text, image, or video) via cross-attention mechanisms to guide the generation process. Recently, LDMs have served as the theoretical backbone of Stable Diffusion (SD) and emerging video generation models such as Sora.
In various industrial applications, LDMs have demonstrated superior performance over GANs~\cite{zhong2023overview}, particularly in defect image generation tasks. To enable the generation of defect images with specific types and locations, AnomalyDiffusion~\cite{hu2024anomalydiffusion} integrates a spatial anomaly embedding module and an adaptive attention weighting mechanism, achieving precise alignment between generated anomalies and ground-truth annotations, thereby significantly enhancing downstream localization performance.
Beyond standard LDMs, fine-tuning techniques such as Textual Inversion~\cite{gal2022image}, Dreambooth~\cite{ruiz2023dreambooth}, LoRA~\cite{hu2022lora}, and InstantID~\cite{wang2024instantid} have been developed to further improve the controllability of SD-based models. However, these approaches typically struggle with accurately controlling the generation location. 
To address this limitation, our model introduces a Targeted Instance-Dependent Spatial Consistency (TIDSC) module, which enforces consistency in the generated location and spatial structure, enabling the acquisition of diverse and realistic defective image–annotation pairs. Furthermore, recent advances in conditional diffusion models, such as progressive conditional diffusion~\cite{shen2023advancing} and rich-contextual conditioning frameworks~\cite{shen2025boosting}, have demonstrated the effectiveness of carefully designed conditioning mechanisms, motivating the design of our TIDSC-enhanced generation pipeline.

\begin{figure*}[t]
\centering
\includegraphics[width=0.9\linewidth]{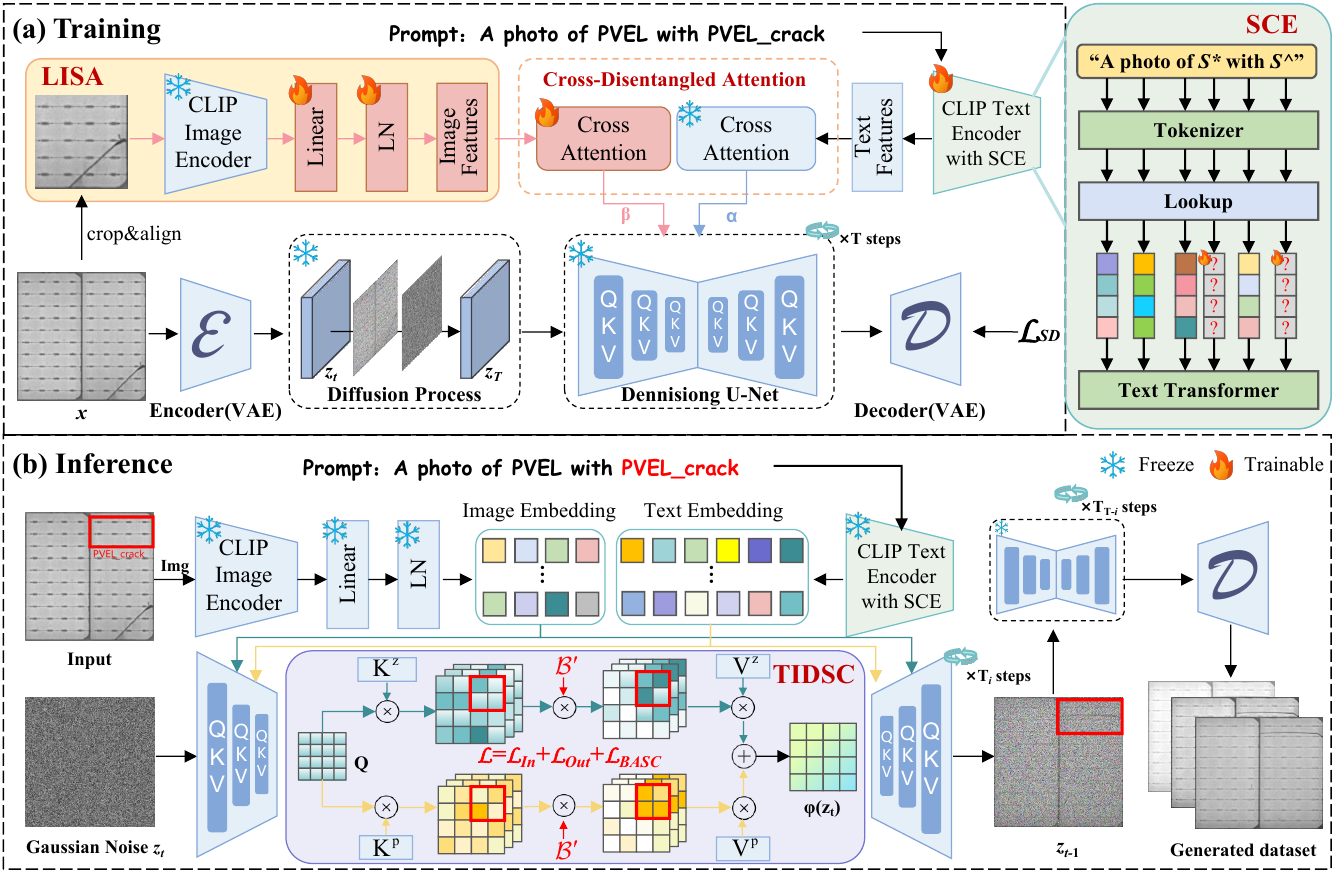}
\caption{The PDIG employs a two-stage generation approach. During the training stage, the Semantic Concept Embedding (SCE) module and the Industrial Style Adaptation (LISA) module learn image and text features, which are then fully integrated into the backbone network to enhance the image generation domain distribution. In the inference stage, the Text-Image Dual-Space Constraints (TIDSC) module is utilized to focus more attention on the specified bounding box regions, thereby enabling precise defect localization generation.}
\label{fig:3}
\end{figure*}
\section{Proposed Method}\label{sec:method} 
In this section, we will detail the proposed PDIG. As shown in Fig. \ref{fig:3}, PDIG employs a two-stage generation approach. During the training stage, the SCE module and the LISA module learn image and text features, which are then fully integrated into the backbone network to enhance the image generation domain distribution. In the inference stage, the TIDSC is utilized to focus more attention on the specified bounding box regions, thereby enabling precise defect localization generation.

\subsection{Preliminaries}
\textbf{Stable Diffusion} model is developed based on the LDM and is trained on the larger dataset LAION-5B. It also employs a more powerful CLIP model to replace the original cross-attention modulation mechanism. Similar to LDM, the SD model consists of three core components:  VAE~\cite{kingma2013auto},  U-Net~\cite{radford2021learning}, and CLIP Text Encoder.
The VAE encoder \( \epsilon \) encodes the image \( x \) into the latent space to obtain \( z_0 \). In the latent space, Gaussian noise is added to \( z_0 \) to obtain \( z_T \). The U-Net network then predicts the noise residual for denoising to obtain \( z \), which is subsequently decoded back to the pixel space by the VAE decoder \( \cal D \) to generate the reconstructed image. The CLIP Text Encoder encodes the input text prompt into a text embedding, which is fed into the cross-attention layer of the U-Net as a conditional control.
The training objective of the model is to minimize the following loss:
\begin{equation}
{{\cal L}_{SD}} = {{\mathbb{E}}_{z \sim \varepsilon (x),\epsilon\sim {\cal N}(0,1),y,t}}\left[ {\left\| {\\ \epsilon - {\\\epsilon_\theta }\left( {{z_t},t,{\tau _\theta }\left( y \right)} \right)} \right\|_2^2} \right],
\end{equation}
Where $t$ is the time step and $y$ is the conditional input, ${\tau _\theta }\left( y \right)$ for the model that maps the conditional input $y$ to the conditional vector.

\textbf{Semantic Concept Embedding} is innovatively introduced by us, embedding industrial-specific concepts into the vocabulary of the CLIP Text Encoder~\cite{gal2022image}. This improvement enables the model to accurately learn text embeddings of industrial concepts and their corresponding visual features, thereby efficiently mastering the unique styles and related semantics of industrial images. We define the text descriptions as ``A photo of ${S^*}$'' and ``A photo of ${S^\wedge }$ '', where ${S^*}$ and ${S^\wedge}$ represent PVEL imaging images and images corresponding to different defect types. To find the optimal embedding vector $v$, our optimization target is:
\begin{equation}
v^* = \mathop {\arg \min }\limits_v {{\mathbb{E}}_{z \sim \varepsilon (x),\epsilon \sim{\cal N}(0,1),y,t}}\left[ {\left\| {\epsilon - {\epsilon_\theta }\left( {{z_t},t,{\tau _\theta }\left( y \right)} \right)} \right\|_2^2} \right].
\end{equation}
${v^*}$ is the marker embedding vector obtained by minimizing the loss of LDM.

\subsection{Lightweight Industrial Style Adaptor (LISA)}  
Due to the lack of more fine-grained learning of image features, existing methods find it difficult to generate industrial defect images with diverse background styles and defect morphologies based solely on text descriptions. To tackle the above issue, we have developed a Lightweight Industrial Style Adaptation (LISA) module. This module adapts the pre-trained SD model by training on limited network levels to achieve the generation of industrial defect images, using image prompts as input. As shown in Fig. \ref{fig:3} (a), PV EL defects are typically small in size, such as scratch, broken gate, and black spot in PV EL defects. Therefore, when processing a defective image \( x \), we first locate the defect regions based on the bounding box information \(\mathcal{B} = \{b_i\}\) provided by defect annotations. Here, \(b_i\) represents the coordinates of the top-left and bottom-right corners of the defect box, i.e., \( \{(x_{i1}, y_{i1}), (x_{i2}, y_{i2})\} \).

\begin{figure}[t]
\centering
\includegraphics[width=0.8\linewidth]{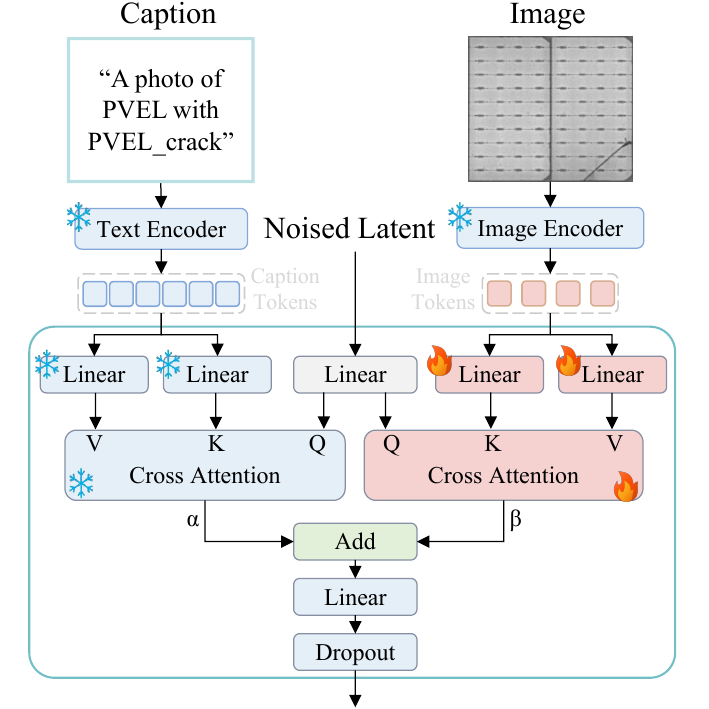}
\caption{Illustration of cross-disentangled attention between Text and Image.}
\label{fig:4}
\end{figure}

Subsequently, we perform cropping and alignment operations on the entire image centered around the box to obtain image feature patches. The purpose of this step is to enable the model to better learn the features of the defect regions. For the extraction of defect image features, we employ a pre-trained CLIP image encoder~\cite{ye2023ip}. After obtaining the global image embedding, we project it through a trainable projection network into a sequence of features with length \(N\), resulting in image tokens \(h_z\). Finally, as shown in Fig. \ref{fig:4}, the image features are integrated in parallel with text features into the denoising UNet module of the SD model via a cross-disentangled attention adaptation module. The attention mechanism can be expressed as follows:
\begin{equation}
{\bar h^z},{\bar h^p} = {\rm{Attn}}([{{\bf{Q}}^z},{{\bf{Q}}^p}],[{{\bf{K}}^z},{{\bf{K}}^p}],[{{\bf{V}}^z},{{\bf{V}}^p}]),\
\end{equation}
Here, \(\bar h^z\) and \(\bar h^p \) represent the outputs of the image and text tokens after cross-attention, respectively. [·,·] denotes concatenation across the tokens dimension; ${{\bf{Q}}^z} = {{\bf{Q}}^p} = {h^p}{\bf{W}}_q^p$, ${{\bf{K}}^z}={h^z}{\bf{W}}_k^z$, ${{\bf{V}}^z} ={h^z}{\bf{W}}_v^z$; ${h_p} = {\tau _\theta }\left( {{y_t}}\right)$;${\bf{W}}_q^p$,${\bf{W}}_k^z$ and  ${\bf{W}}_v^z$ are the weight matrices of the linear projection layers for the query of text tokens, and the key and value of image tokens, respectively. Based on the attention mechanism, the final formula for cross-attention between image and text can be derived as follows:

\begin{equation}
\begin{aligned}
& \bar H 
= \alpha  \cdot {{\bar h}^z} + \beta  \cdot {{\bar h}^p}\\
&= \alpha  \cdot {\mathop{\rm Attn}\nolimits} ({{\bf{Q}}^p},{{\bf{K}}^z},{{\bf{V}}^z}) + \beta  \cdot {\mathop{\rm Attn}\nolimits} ({{\bf{Q}}^p},{{\bf{K}}^p},{{\bf{V}}^p})\\
&= \alpha  \cdot {\mathop{\rm Softmax}\nolimits} (\frac{{{{\bf{Q}}^p}{{({{\bf{K}}^z})}^ \top }}}{{\sqrt d }}){{\bf{V}}^z}
+\beta \cdot {\mathop{\rm Softmax}\nolimits}(\frac{{{{\bf{Q}}^p}{{({{\bf{K}}^p})}^ \top }}}{{\sqrt d }}){{\bf{V}}^p.}
\end{aligned}
\end{equation}
Here, $\alpha$ and $\beta$ are weighting factors that can adjust the proportion of text and image prompts.

\subsection{Text-Image Dual-Space Constraints (TIDSC)}  
While large-scale data generation requires a corresponding volume of annotations, the efficiency of manual annotation can no longer meet the actual needs. Thus, to realize defect location generation and semi-automatic annotation, we propose the TIDSC module. In the inference stage, the user provides the caption prompt, the image prompt, and the location of the defect to be generated ${\cal B}'=\left\{{{b_j}}\right\}$. From the LISA module, we can get a set of caption tokens and image tokens ${\cal H} = \left\{ {h_j^p,h_j^z} \right\}$ as well as cross attention maps ${\cal \bar H}^t =\{ \bar H_j^t\} $. 
Further, according to a given ${\cal B}'$, which gives a set of binary spatial masks ${\cal M}=\{{M_j}\}$. Our goal is to generate target objects in the mask area whenever possible~\cite{xie2023boxdiff}.
To achieve this goal, we propose graphic bi-spatial constraints on the target cross-attention map, including masked internal and external constraints, as well as boundary alignment smoothing constraints. These constraints gradually update the underlying  \( z_t \) so that the position and scale of the composite object agree with the mask region.
For internal constraints, our goal is to constrain responses with high attention values within the masked region, defined as:
\begin{equation}
{\cal L}_{{H_j}}^1 = 1 - \frac{1}{P}\sum {{\bf{topk}}} \left( {\bar H_j^t \cdot {{\bf{M}}_j},P} \right),
\end{equation}
\begin{equation}
{{\cal L}_{In}} = \sum\limits_{{H_j} \in {\cal H}} {{\cal L}_{{H_j}}^1},
\end{equation}
where means that \( P \) elements with the highest response would be selected.

Internal constraints guarantee that the image of the target image is generated in the masked area, but not how the image is generated outside the area, so for external constraints, we need the attention response as low as possible, therefore:
\begin{equation}
{\cal L}_{{H_j}}^2 = \frac{1}{P}\sum {{\bf{topk}}} \left( {\bar H_j^t \cdot (1 - {{\bf{M}}_j}),P} \right),
\end{equation}
\begin{equation}
{{\cal L}_{Out}} = \sum\limits_{{H_i} \in {\cal H}} {{\cal L}_{{H_j}}^2} ,
\end{equation}

To obtain precise object or context boundary pixels to restrict the scale while ensuring that the boundaries of objects in the generated images are smoother and more natural, we propose the Boundary Smoothing Alignment Constraint. First, to maintain the position and scale consistency of objects within the specified bounding boxes, we project each target mask \( M_j \) and the cross-attention map onto the x-axis and y-axis via the max operation. For the x-axis:
\begin{equation}
{{\bf{m}}_x}(k) = {\bf{ma}}{{\bf{x}}_{s = 1, \cdots ,S}}\{ {{\bf{M}}_j}(s,k)\},
\end{equation}
\begin{equation}
{\bf{a}}_x^t(k) = {\bf{ma}}{{\bf{x}}_{s = 1, \cdots ,S}}\{ \bar H_j^t(s,k)\},
\end{equation}
Among them,${{\bf{m}}_x}(k),{\bf{a}}_x^t(k) \in {{\mathbb R}^K}$. Our goal is to optimize ${\bf{a}}_x^t$ so that it is close to ${{\bf{m}}_x}$:
\begin{equation}
{\cal L}_{{H_j}}^3 = \frac{1}{L}\sum {\bf{tokp}}\left( {\{ |{{\bf{m}}_x}(k) - {\bf{a}}_x^t(k)|\} _{k = 1}^K,L,x_1^j,x_2^j} \right),
\end{equation}
Here, ${\rm{tokp}}( \cdot ,L,x_1^j,x_2^j)$ denotes the uniform sampling of \( L \) error terms from $\{ |{{\bf{m}}_x}(k) - {\bf{a}}_x^t(k)|\} _{k = 1}^K$, with these error terms surrounding the given alignment coordinates $x_1^j,x_2^j$ on the x-axis.

Apply the same operation to the y-axis:
\begin{equation}
{{\bf{m}}_y}(j) = {\bf{ma}}{{\bf{x}}_{k = 1, \cdots ,K}}\{ {{\bf{M}}_j}(s,k)\},
\end{equation}
\begin{equation}
{\bf{a}}_y^t(j) = {\bf{ma}}{{\bf{x}}_{k = 1, \cdots ,K}}\{ \bar H_j^t(s,k)\},
\end{equation}
\begin{equation}
{\cal L}_{{H_j}}^4 = \frac{1}{L}\sum {\bf{tokp}}\left( {\{ |{{\bf{m}}_y}(j) - {\bf{a}}_y^t(j)|\} _{j = 1}^S,L,y_1^j,y_2^j} \right),
\end{equation}

To ensure that the boundaries of objects in the generated images are smoother and more natural, we achieve the smoothing effect by minimizing the gradient variations and second-order derivatives of the cross-attention maps:
\begin{equation}
{\cal L}_{{\rm sm}_j}^{(1)} =\frac{1}{{S \times K}} \sum\limits_{(x,y)\in{M_j}} {\left( {\left| {\frac{{\partial \bar H_j^t(x,y)}}{{\partial x}}} \right| + \left| {\frac{{\partial \bar H_j^t(x,y)}}{{\partial y}}} \right|} \right)},
\end{equation}
\begin{equation}
{\cal L}_{{\rm sm}_j}^{(2)} =\frac{1}{{S \times K}} \sum\limits_{(x,y)\in{M_j}} {\left( {\left| {\frac{{{\partial ^2}\bar H_j^t(x,y)}}{{{\partial ^2}x}}} \right| + \left| {\frac{{{\partial ^2}\bar H_j^t(x,y)}}{{{\partial ^2}y}}} \right|} \right)},
\end{equation}

The boundary smoothing alignment constraint is:
\begin{equation}
{{\cal L}_{BASC}} = \sum\limits_{{H_j} \in {\cal H}} {{\rm{(}}{\cal L}_{{H_j}}^3 + {\cal L}_{{H_j}}^4)}  + {\lambda _1}\sum{\cal L}_{{\rm sm}_j}^{(1)} + {\lambda _2}\sum{\cal L}_{{\rm sm}_j}^{(2)},
\end{equation}
Where \( \lambda _1\) and \(\lambda _2\) are the smoothing weights for the first-order and second-order terms, respectively.

Therefore, our total constraints are follows:
\begin{equation}
{\cal L} = {{\cal L}_{In}} + {{\cal L}_{Out}} + {{\cal L}_{BASC}}.
\end{equation}

Through the graphic dual space restriction module, we asked the model to allocate more attention to the specified bounding box area, so as to realize defect positioning and generation, and output image annotation files to provide annotation information for downstream applications.

\section{Experiment and Analysis}\label{sec:exp}  
To demonstrate the superiority of the proposed PDIG, we compare it with multiple state-of-the-art image generation approaches on the ELES dataset.
\begin{table}[t] \tiny 
	\centering
	\caption{ The initial dataset allocations for training and testing}
	\label{tab:t1}
	\setlength{\tabcolsep}{1pt}{
		\renewcommand\arraystretch{1.2}
		\begin{tabular}{ccccccccc}
			\hline
			\multirow{2}{*}{Datasets} & \multicolumn{2}{c}{EL group1} &  & \multicolumn{2}{c}{EL group2} &  & \multicolumn{2}{c}{EL group3} \\ \cline{2-3} \cline{5-6} \cline{8-9} 
			& defective image    & defect-free image   &  & defective image    & defect-free image   &  & defective image    & defect-free image   \\ \hline 
			Training                  & 6652               & $\backslash$        &  & $\backslash$       & $\backslash$        &  & $\backslash$       & $\backslash$        \\
			Testing                   & 1110               & 1410                &  & 3209               & 2541                &  &839                & 562                  \\
			Total                     & \multicolumn{2}{c}{9172}                 &  & \multicolumn{2}{c}{5750}                 &  & \multicolumn{2}{c}{1401}                 \\
			Acquisition time          & \multicolumn{2}{c}{September 2020}       &  & \multicolumn{2}{c}{May 2021}             &  & \multicolumn{2}{c}{October 2021}         \\
			Resolution                & \multicolumn{2}{c}{$384\times384$}       &  & \multicolumn{2}{c}{$640\times589$}       &  & \multicolumn{2}{c}{$700\times668$}       \\ \hline
		\end{tabular}
	}
	
\end{table}
\begin{figure}[t]
\centering
\includegraphics[width=0.7\linewidth]{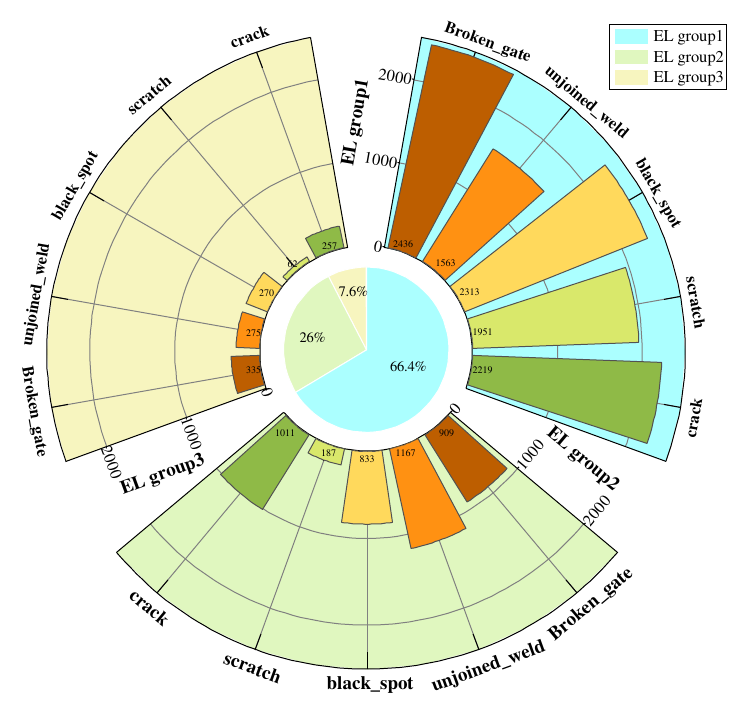}
\caption{Distribution of defect categories in each production line of the dataset. It illustrates the significant differences in the proportion of data among the three groups and the distribution of defect quantities within each group. For example, the number of ``scratch" defects in EL group3 is only 62, highlighting a noticeable numerical discrepancy.}
\label{fig:5}
\end{figure}
\subsection{Datasets}
\textbf{\emph{ELES Dataset}}~\cite{zhao2023ssn} is the first PV module EL endogenous displacement dataset, created from three sets of EL images collected at different times, a total of 16,323. For a fair comparison, the training and testing sets of this study were assigned according to the literature~\cite{zhao2023ssn}. As shown in Table  \ref{tab:t1}, we only used EL group1 to train the PDIG model. Fig. \ref{fig:5} depicts the distribution of defect categories for the ELES dataset. At the same time, in order to realize the multimodal data input of the model, we improve the text description of the dataset according to each image defect type, which is defined as ``A photo of \{\} with \{\} and \{\}".

\subsection{Evaluation Metrics}  
In the stage of generating image quality assessment, we use Frechet Inception Distance (FID)~\cite{heusel2017gans} and Inception Score (IS)~\cite{barratt2018note} as the main evaluation indicators, and the FID is defined as:

\begin{small}
\begin{equation}
FID\left( {x,\tilde x} \right) = \left\| {{u_x} - {u_{\tilde x}}} \right\|_2^2 + Tr\left( {{\sum _x} + {\sum _{\tilde x}} - 2{{\left( {{\sum _x}{\sum _{\tilde x}}} \right)}^{0.5}}} \right),
\end{equation}
\end{small}
$Tr$ denotes the composite of the elements on the diagonal of the matrix, which becomes the trace of the matrix in matrix theory. $x$ and $\tilde x$ denote the real and generated images, $u$ denotes the mean, $\sigma$ is the covariance matrix. A lower FID means that the generated image performs better in terms of fidelity and diversity.

IS evaluates the performance of the model by calculating the KL scatter of the predictive probability distribution of the categories of the generated images, and IS is defined as:

\begin{equation}
IS = \exp \left( {{{\mathbb{E}}_x}\left[ {{\rm{KL}}(p(y|x)||p(y))} \right]} \right).
\end{equation}
Where ${p(y|x)}$ is the distribution of category prediction probabilities for a given image $x$, ${p(y)}$ is the average distribution of category prediction probabilities for all generated images, and higher IS values indicate that the generated images are of high quality and diversity.

\begin{figure*}[t]
\centering
\includegraphics[width=0.98\linewidth]{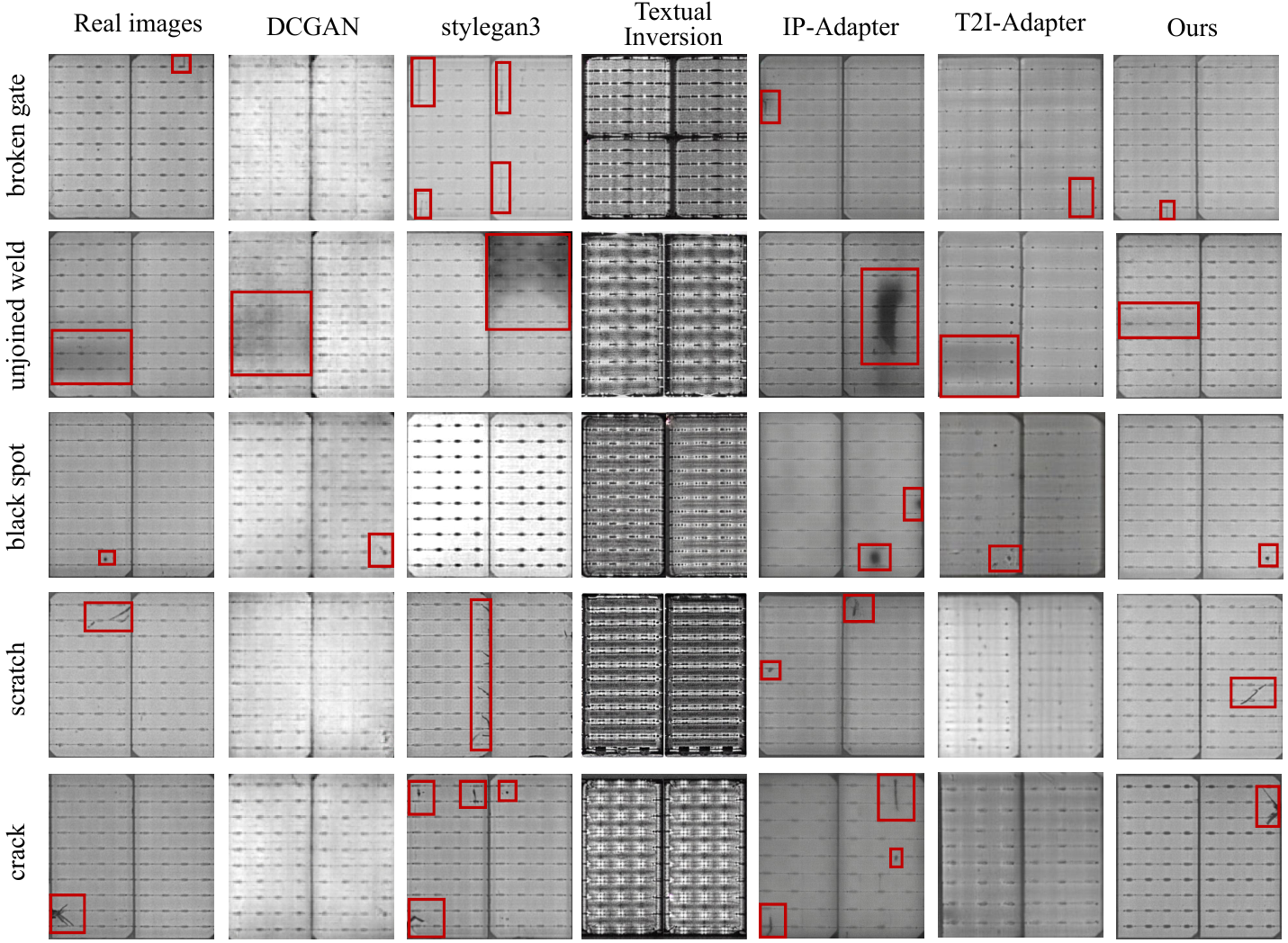}
\caption{Qualitative comparison of ours with other models. Generated target defects are highlighted in red boxes.}
\label{fig:6}
\end{figure*}
\subsection{Implementation Details} 
We trained the network using Python 3.10 on an NVIDIA RTX 3090 GPU. For the pre-trained model, we selected the lightweight and efficient SDv1.5 version and used the high-accuracy, multi-modal CLIP ViT-H/14 - LAION-2B model as the image encoder. During the training stage, we trained the proposed model using the MSE loss, with AdamW as the optimizer. The initial learning rate was set to 0.0001, weight decay to 0.01, and the generated image resolution was set to 512×512. The batch size was set to 8, and the training was conducted for over 1000 epochs.
In the inference stage, we employed the DDIM sampler with the initial classifier guidance scale set to 7.5. When using only image prompts, we set $\alpha=0$ and $\beta=1.0$. Additionally, we set the inference steps to 60 and applied the TIDSC mechanism only during the first 30 steps.
\begin{table}[t]
\scriptsize
\centering
\caption{Comparison of different image generation methods for various defect types. \textbf{Bold} results represent the optimal generation performance. \underline{underline} results represent the second optimal generation performance. }
\label{tab:t2}
\setlength{\tabcolsep}{0.8pt}{
    \renewcommand\arraystretch{1.2}
    \begin{tabular}{c>{\centering\arraybackslash}p{0.7cm}>{\centering\arraybackslash}p{0.9cm}>{\centering\arraybackslash}p{0.95cm}>{\centering\arraybackslash}p{1.2cm}>{\centering\arraybackslash}p{1.1cm}>{\centering\arraybackslash}p{1.2cm}c}
    \toprule
     Defect types &  Metric& DCGAN\newline\cite{patil2021dcgan} & Stylegan3\newline\cite{Karras2021} & Textual\newline Inversion\cite{gal2022image}&  IP-adapter\newline\cite{ye2023ip} &T2I-adapter\newline\cite{mou2024t2i} &Ours\\
    \midrule
    \multirow{2}{*}{broken gate}& FID $\downarrow$ & 151.97 & \underline{39.6} & 175.90 & 58.56 & 73.41 & \textbf{22.37} \\
     & IS $\uparrow$ & 1.29 & 1.29 & \underline{}{1.62} & 1.51 & 1.56 & \textbf{1.65} \\
    \multirow{2}{*}{unjoined weld}& FID $\downarrow$ & 140.70 & \underline{35.5} & 171.23 & 56.62 & 76.37 & \textbf{19.03} \\
     & IS $\uparrow$ & 1.26 & 1.36 & \textbf{1.99} & 1.38 & 1.41 & \underline{1.48} \\
    \multirow{2}{*}{black spot}& FID $\downarrow$ & 150.03 & 51.32 & 231.30 & \underline{32.08} & 104.15 & \textbf{20.56} \\
     & IS $\uparrow$ & 1.26 & 1.34 & \textbf{2.68} & 1.38 & 1.52 & \underline{1.54} \\
    \multirow{2}{*}{scratch} & FID $\downarrow$ & 149.21 & \underline{34.9} & 219.32 & 39.34 & 96.14 & \textbf{23.20} \\
     & IS $\uparrow$ & 1.24 & 1.26 & \textbf{2.70} & 1.42 & 1.41 & \underline{1.50} \\
     \multirow{2}{*}{crack} & FID $\downarrow$ & 152.38 & \underline{39.00} & 187.42 & 59.07 & 88.85 & \textbf{19.34} \\
     & IS $\uparrow$ & 1.26 & 1.28 & \textbf{2.40} & 1.38 & \underline{1.56} & 1.46 \\
    \midrule
    \multirow{2}{*}{Average} & FID $\downarrow$ &148.86 & \underline{40.06} & 197.03 & 49.13 & 87.78 & \textbf{20.90} \\
     & IS $\uparrow$ & 1.26 & 1.32 & \textbf{2.28} & 1.41 & 1.49 & \underline{1.53} \\
    \bottomrule
    \end{tabular}}
\end{table}
\subsection{Comparison With Recent Works} 
In this section, we selected several state-of-the-art image-generation algorithms currently used in the industrial field, including the GAN series, such as DCGAN~\cite{patil2021dcgan} and StyleGAN3~\cite{Karras2021}, as well as text-to-image methods like Textual Inversion~\cite{gal2022image}, IP-adapter~\cite{ye2023ip}, and T2I-adapter~\cite{mou2024t2i}. We evaluated the effectiveness of our method from both qualitative and quantitative perspectives.

We trained our model on the training set allocated for EL group1 images, as shown in Table \ref{tab:t1}. During the inference stage, we generated 200 defect images for each type of defect to calculate the FID and IS metrics. The quantitative results of the models are presented in Table \ref{tab:t2}. For various types of industrial defects, our PDIG method significantly outperforms other existing methods in terms of the FID. This result strongly demonstrates the effectiveness of the introduced SCE module and the proposed LISA module. These modules are capable of fully learning the industrial defect features embedded in images and text, including background styles and fine-grained features of different types of defects, and generating high-quality and diverse defect data.
The average FID value of our method is 19.16 lower than that of the second-best StyleGAN3 method. In terms of the IS metric, our method ranks second only to Textual Inversion. However, Textual Inversion has an FID value that is 176.13 higher than ours. This indicates that the images generated by Textual Inversion are produced at the expense of extremely low fidelity, in order to achieve higher diversity. Such a trade-off is evidently meaningless.

The qualitative results of our model compared to other methods are shown in Fig. \ref{fig:6}. Among the various types of defect images, the backgrounds generated by DCGAN are relatively blurry, and the effectiveness of defect generation is relatively low. StyleGAN3 can generally handle the background of defect images well, but it struggles to accurately interpret smaller defects such as black spots. Among text-to-image methods, Textual Inversion performs the worst. This is attributed to its training on only a limited number of images to learn the style, without adequately addressing the detailed features of the images. Consequently, the generated images exhibit significant differences from the original images. This also explains why Textual Inversion achieves the highest IS in Table \ref{tab:t2}, as the increased diversity comes at the cost of authenticity.
The IP-Adapter shows better performance in background generation but exhibits lower accuracy in generating individual defect features. For instance, it tends to produce black spots when generating crack images. The T2I-Adapter method, while effective in some aspects, falls short in background generation compared to the IP-Adapter, with issues such as misaligned and disconnected grid lines. Additionally, it generates defects with poor quality and abnormal textures.
\begin{figure}[t]
\centering
\includegraphics[width=1\linewidth]{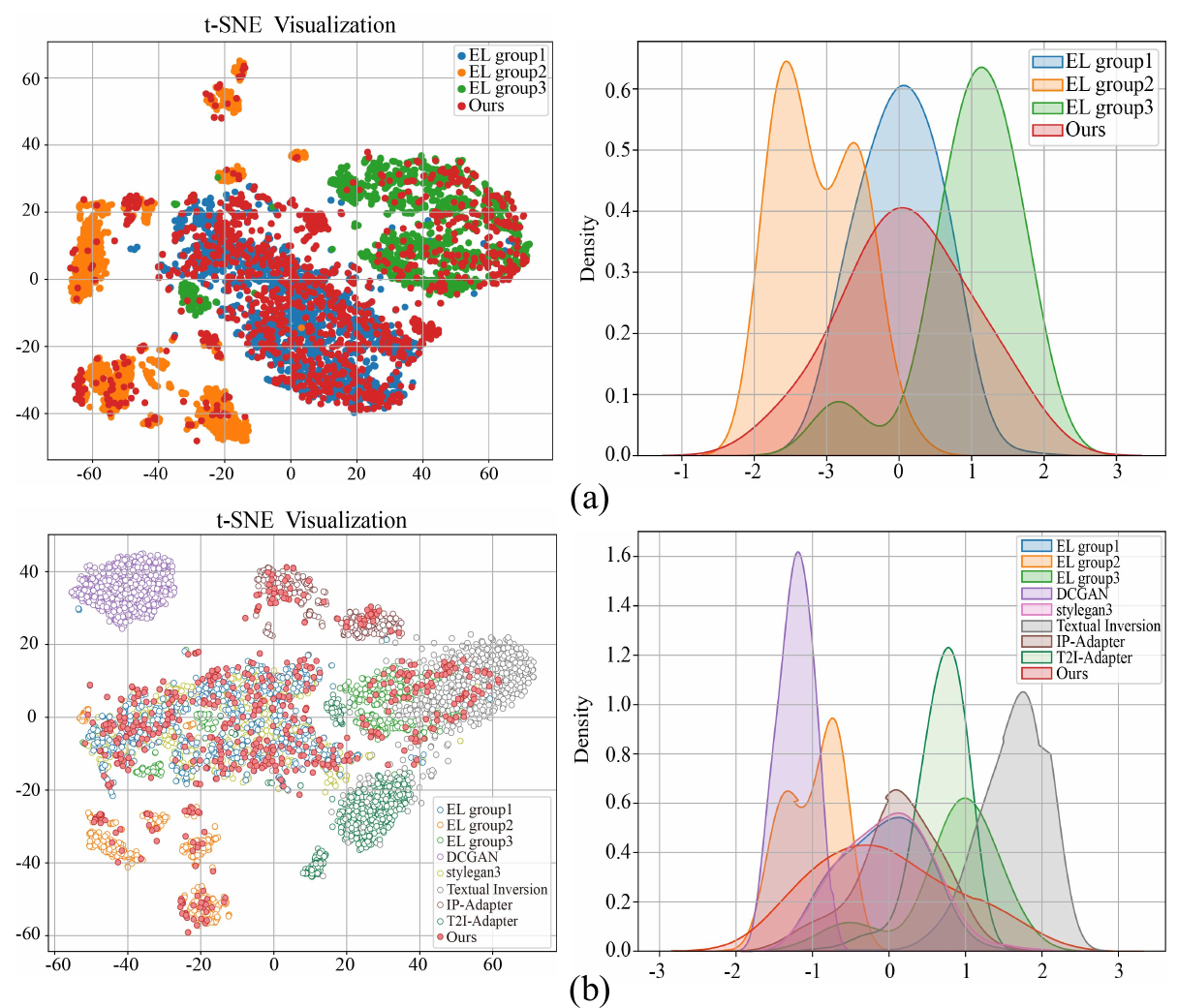}
\caption{Comparison of t-SNE Dimensionality Reduction and Probability Distribution Visualization for Generated Images: (a) Comparison of images generated using our method with the three datasets in ELES. (b) Comparison with images generated using different generation methods.}
\label{fig:7}
\end{figure}
\begin{figure}[t]
\centering
\includegraphics[width=1\linewidth]{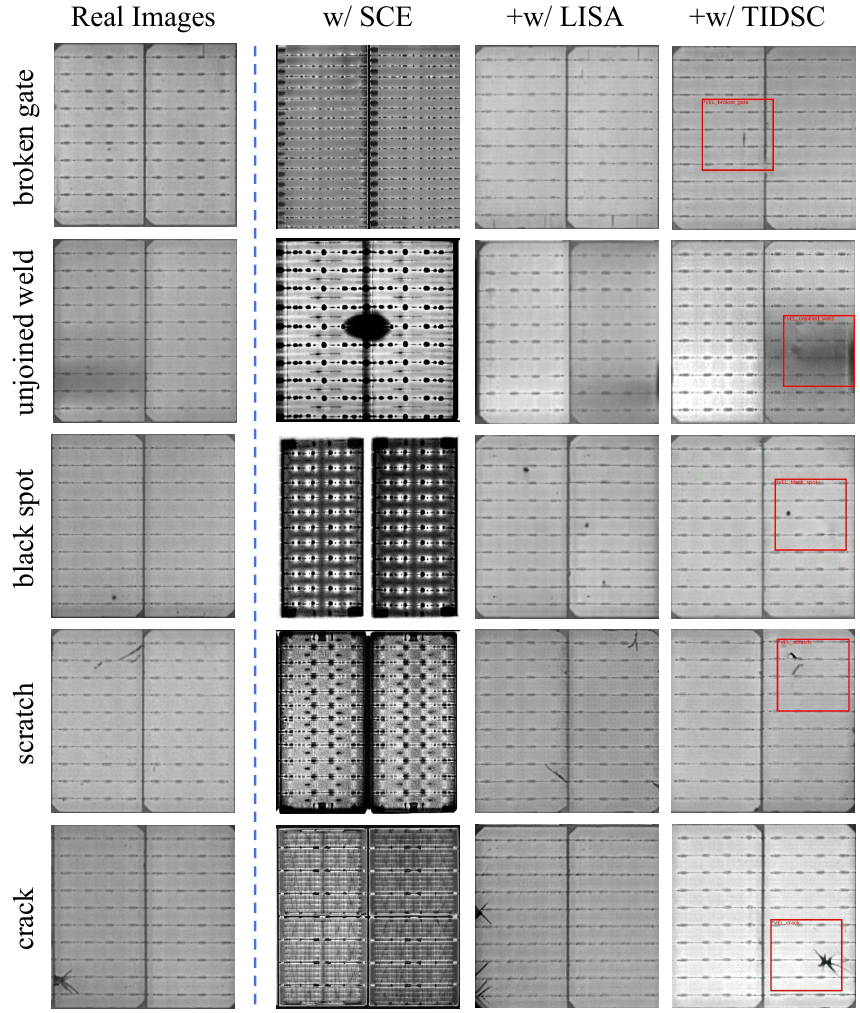}
\caption{Visualization of ablation experiments. The results generated at each stage for different defect types are shown. The user-specified box is marked in red.}
\label{fig:8}
\end{figure}
\begin{table}[t]
\centering
\caption{Ablation study results. \textbf{Bold} results represent the optimal generation performance. }
\label{tab:t3}
\begin{tabular}{cc|c|cc}
\toprule
\multicolumn{2}{c|}{\textbf{Training}} &  \textbf{Inference} &\multicolumn{2}{c}{\textbf{Metric} } \\
\midrule
\textbf{SCE}& \textbf{LISA} & \textbf{TIDSC} & \textbf{FID $\downarrow$} & \textbf{IS $\uparrow$} \\
\midrule
\checkmark & \texttimes &\texttimes &197.03&2.28\\
\texttimes & \checkmark &\texttimes &48.64 &1.39\\
\texttimes & \texttimes &\checkmark &265.31&\textbf{3.72} \\
\checkmark & \checkmark &\texttimes &36.78 &1.40 \\
\checkmark & \texttimes &\checkmark &174.13&2.05 \\
\texttimes & \checkmark &\checkmark &40.22 &1.42 \\
\rowcolor{gray!20}
\checkmark & \checkmark &\checkmark &\textbf{20.90} & 1.53 \\
\bottomrule
\end{tabular}
\end{table}
In contrast, our method achieves results that are highly consistent with real defect images. It demonstrates high-quality and authentic generation in both background and defect morphology, effectively capturing the essential features of the defects. Meanwhile, thanks to the strong denoising capability of the diffusion model, our method can generate data with a rich and diverse distribution. Fig. \ref{fig:7} (a) compares the image distribution generated by our method and the original dataset distribution. It can be seen that the generated images can cover all the dataset distribution and reduce the problem of endogenous bias. Fig. \ref{fig:7} (b) compares the image distribution generated by different methods.

\subsection{Ablation Studies and Analysis} 
In this section, we conduct an ablation study using the model trained on the EL group1 dataset to evaluate the impact of each component of our method. The results of this analysis are presented in Table \ref{tab:t3}, which demonstrates the contributions of the individual components to the overall performance.

\emph{1) Contribution of SCE:} We have evaluated the impact of the SCE module on generation. As shown in Fig. \ref{fig:8} and Table \ref{tab:t3}, we can observe that this module is capable of effectively generating the style of PV EL images and thoroughly learning the structural features of PV grid lines. However, the module only performs a rough learning of the style and embeds defect-related text into the model. It is still not able to effectively distinguish and generate smaller-sized defects.

\emph{2) Contribution of LISA:} We have assessed the impact of the LISA module on image generation. As shown in Table \ref{tab:t3}, when only using the LISA module, the FID score is 48.64 and the IS score is 1.39. Compared to adding only the SCE module, the FID has decreased by 148.39, indicating a significant improvement in the quality and realism of the generated images. However, the IS has dropped by 0.89, which is attributed to the generated images being more similar to real images, thereby reducing image diversity.
When the LISA module is added based on the SCE module, the FID is further reduced by 11.86. As shown in Fig. \ref{fig:8}, after the LISA module is incorporated, the generated defect images are almost identical to the original images. The module is capable of effectively learning the features of various defects and can meticulously handle small-target defects, such as broken gate and black spot. For the unjoined weld defect, it can adeptly manage the contrast between the defect and the background. We have determined that the LISA module contains a total of \textbf{22.32M} trainable parameters. Specifically, the linear projection layer contributes \textbf{3.15M} parameters, while the cross-attention layers account for \textbf{19.17M} parameters.

\emph{3) Contribution of TIDSC:} We have evaluated the impact of the TIDSC module on image generation. During the inference stage, we incorporate the TIDSC module to achieve location-specific generation while simultaneously pre-generating annotations. As shown in Table \ref{tab:t3}, when the TIDSC module is used alone, it essentially imposes text boundary control on the SD model. At this point, no feature learning of PV EL images takes place. Therefore, the metrics obtained from images generated solely based on text prompts are as follows: an FID of 265.31 and an IS of 3.72. Although the IS metric is the best in this case, it is meaningless to trade off image realism for diversity.
\begin{figure}[t]
\centering
\includegraphics[width=0.9\linewidth]{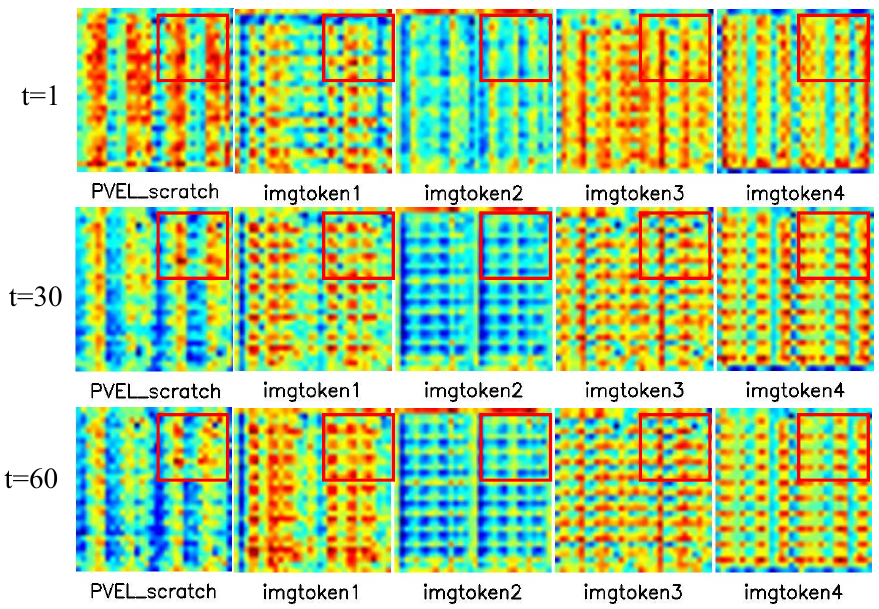}
\caption{Visualization Results of the TIDSC Module. For different selected moments, the attention maps corresponding to the tokens describing the defect types and four image tokens are shown. The red box indicates the predefined regions of interest.}
\label{fig:9}
\end{figure}
Furthermore, when we combine the SCE with the TIDSC module, the FID decreases by 91.18 and the IS decreases by 0.23. When LISA and TIDSC are combined, there is a noticeable improvement in the quality of generated images, with an FID of 40.22. Finally, when all three modules are used together, the FID of the generated images is 20.9, and the IS is 1.53. This demonstrates that our method produces images with high realism and good diversity.

As shown in Fig. \ref{fig:8}, our method can generate specified defect images within the given bounding box and effectively produces images with diverse defect shapes and backgrounds. Fig. \ref{fig:9} shows the visualization results of the TIDSC module. When t=1, the attention values of the text and image tokens are uniformly distributed. When t=30, based on the text prompt, the attention can be focused as much as possible within the box. While focusing on the box, the image tokens pay more attention to the overall detailed features of the image.

The limitation is that the TIDSC module has not yet achieved absolute restriction. Due to the varying sizes and positions of defects, the generated defects may overflow, as seen in the unjoined weld and crack images in Fig. \ref{fig:8}. To verify the pass rate of image localization generation and the bounding box overflow situation, we generated 200 images for each type of defect data in batches and manually annotated the qualified generated images to test the effectiveness of the module. The verification results are shown in Table \ref{tab:t4}, with an average image generation pass rate of 84.74$\%$ and an average box spillover rate of 8.37$\%$. This indicates a relatively good effect on defect localization generation and provides a way to achieve automatic annotation for generated images.

\begin{table}[t]
\centering
\caption{The pass rate of image positioning generation and the spillover rate of box.}
\label{tab:t4}
\begin{tabular}{ccc}
\toprule
\textbf{Defect types}& \textbf{Pass Rate $\%$} &\textbf{Box Spillover Rate $\%$}\\
\midrule
broken gate  &81.90 &8.70\\
unjoined weld &83.80 &10.41\\
black spot    &88.70 &4.35\\
crack &85.30 &9.00\\
scratch &84.00&9.40\\
\midrule
\textbf{average} &\textbf{84.74} &\textbf{8.37} \\
\bottomrule
\end{tabular}
\end{table}

\begin{table}[t]
\scriptsize
\centering
\caption{Verify the performance of adding our generated data to the yolo models for defect detection. \textcolor{green}{$\downarrow$} indicates a decrease in the indicator. \textcolor{red}{$\uparrow$} indicates an increase in the indicator. }
\label{tab:t5}
\setlength{\tabcolsep}{0.75pt}{
    \renewcommand\arraystretch{1.2}
    \begin{tabular}{cc|ccccc|c}
    \toprule
     Datasets & methods & broken gate &unjoined weld & black spot & scratch &crack &Average\\
    \midrule
    \multirow{4}{*}{EL group1} 
     & yolov5 &91.7  &85.5  &68.2  &73.8  & 80.4 &79.9 \\
     & +ours  & 88.4$_{\textcolor{green}{\downarrow3.3}}$ & 86.4$_{\textcolor{red}{\uparrow0.9}}$ & 70.6$_{\textcolor{red}{\uparrow2.7}}$ & 70.9$_{\textcolor{red}{\downarrow2.9}}$ & 88.6$_{\textcolor{red}{\uparrow8.2}}$ &81$_{\textcolor{red}{\uparrow1.1}}$\\
     & yolov8 & 95.5 & 89.1 & 76.7 & 78.3 & 93.4 & 86.6 \\
     & +ours & 88.9$_{\textcolor{green}{\downarrow6.6}}$ & 93.1$_{\textcolor{red}{\uparrow3.4}}$ & 79.2$_{\textcolor{red}{\uparrow2.5}}$ & 80.8$_{\textcolor{red}{\uparrow2.5}}$ & 94.4$_{\textcolor{red}{\uparrow1}}$ &87.3$_{\textcolor{red}{\uparrow0.7}}$ \\
     \midrule
    \multirow{4}{*}{EL group2} 
     & yolov5 &65.4 &76.7 &59.5 &55.4 &31.2 &57.6 \\
     & +ours  &69.5$_{\textcolor{red}{\uparrow6.6}}$ & 82.4$_{\textcolor{red}{\uparrow5.7}}$ & 60.9$_{\textcolor{red}{\uparrow1.4}}$ & 60.5$_{\textcolor{red}{\uparrow5.1}}$ & 43.61$_{\textcolor{red}{\uparrow12.41}}$ &63.4$_{\textcolor{red}{\uparrow5.8}}$\\
     & yolov8 &67.3 &79.9 &59.9 &62.1 &48.7 &63.6 \\
     & +ours  &70.1$_{\textcolor{red}{\uparrow2.8}}$ & 81.7$_{\textcolor{red}{\uparrow1.8}}$ & 58.5$_{\textcolor{green}{\downarrow1.4}}$ & 65.7$_{\textcolor{red}{\uparrow3.6}}$ & 50.9$_{\textcolor{red}{\uparrow2.2}}$ &65.4$_{\textcolor{red}{\uparrow1.8}}$\\
     \midrule
     \multirow{4}{*}{EL group3} 
     &yolov5  &57.5 &40.3 &47.3 &34 &31.7 &42.2 \\
     & +ours  &61.8$_{\textcolor{red}{\uparrow4.3}}$ & 38.9$_{\textcolor{green}{\downarrow1.4}}$ & 48.9$_{\textcolor{red}{\uparrow1.6}}$ & 40.8$_{\textcolor{red}{\uparrow6.8}}$ & 52$_{\textcolor{red}{\uparrow20.3}}$ &48.5$_{\textcolor{red}{\uparrow6.3}}$\\
     & yolov8 &58 &39.5 &44.6 &40.4 &48.8 &46.3 \\
     & +ours  &60.7$_{\textcolor{red}{\uparrow2.7}}$ & 42.4$_{\textcolor{red}{\uparrow2.9}}$ & 46.2$_{\textcolor{red}{\uparrow1.6}}$ & 43.3$_{\textcolor{red}{\uparrow2.9}}$ & 48.2$_{\textcolor{green}{\downarrow0.6}}$ &48.1$_{\textcolor{red}{\uparrow1.8}}$\\
    \bottomrule
    \end{tabular}}
\end{table}

\subsection{Defect Detection} 
To verify that the images generated using our method can enhance the performance of defect detection tasks, we generated 200 images and corresponding annotations for each type of defect sample. Currently, for the identification of small target defects, we employed the effective YOLOv5 and YOLOv8 algorithms. We trained the aforementioned detection algorithms on the EL group1 dataset and the expanded EL group1 dataset, and then conducted defect detection on the test set of EL group1, as well as on EL group2 and EL group3. 

The experimental results are shown in Table \ref{tab:t5}. After incorporating the generated samples, the average mAP of the detection model has improved. When our generated images were added to YOLOv5, the average mAP of the model increased by 5.8$\%$ and 6.3$\%$ on EL group2 and EL group3, respectively. Based on YOLOv8, our model achieved an average mAP improvement of 1.8 on both EL group2 and EL group3. These results confirm that our method, by generating diverse images, can effectively mitigate the low detection accuracy caused by endogenous shifts.

\section{Conclusion}\label{sec:con} 
This paper proposed the PDIG, which focuses on the endogenous shifts in real PV scenarios, the poor consistency and controllability of generated images, and the lack of labels for the data. First, PDIG used a few industrial defect images and added embedded priors through the SCE module to capture the specific relational concepts between defect types and their appearances. Next, we employed the LISA module to embed industrial image defect features into the SD model using a cross-disentangled attention manner, thereby enhancing the image generation domain distribution. Finally, during the inference stage, we introduced the TIDSC module to ensure the quality of generated images via positional consistency and spatial smoothing alignment. Extensive experiments have shown that our model significantly outperforms state-of-the-art methods in terms of both realism and diversity of generation and effectively improves the performance of downstream defect object detection tasks. However, our method also has limitations, such as image localization generation overflow. In the future, we will focus on further improving the accuracy of image localization generation and addressing the issue of generating multiple instances simultaneously.

\bibliographystyle{IEEEtran}
\bibliography{main}

\end{document}